# UniPTMs: The First Unified Multi-type PTM Site Prediction Model via Master-Slave Architecture-Based Multi-Stage Fusion Strategy and Hierarchical Contrastive Loss

Yiyu Lin[1], Yan Wang[3,4], You Zhou[3], Xinye Ni[2], Jiahui Wu[1], Sen Yang[1,2,*]

[1]School of Computer Science and Artificial Intelligence Aliyun School of Big Data School of Software, Changzhou University, Changzhou 213164, China.

[2]Changzhou Second People's Hospital, The Third Affiliated Hospital of Nanjing Medical University, Changzhou Medical Center, Nanjing Medical University, Changzhou, 213003, China, Changzhou 213164, China.

[3]Key Laboratory of Symbol Computation and Knowledge Engineering of Ministry of Education, and College of Computer Science and Technology, Jilin University, Changchun, China

[4]School of Artificial Intelligence, Jilin University, Changchun 130012, China

*Correspondence: ys@cczu.edu.cn (Sen Yang)

## Abstract

As a core mechanism of epigenetic regulation in eukaryotes, protein post-translational modifications (PTMs) require precise prediction to decipher dynamic life activity networks. To address the limitations of existing deep learning models in cross-modal feature fusion, domain generalization, and architectural optimization, this study proposes UniPTMs: the first unified framework for multi-type PTM prediction. The framework innovatively establishes a "Master-Slave" dual-path collaborative architecture: The master path dynamically integrates high-dimensional representations of protein sequences, structures, and evolutionary information through a Bidirectional Gated Cross-Attention (BGCA) module, while the slave path optimizes feature discrepancies and recalibration between structural and traditional features using a Low-Dimensional Fusion Network (LDFN). Complemented by a Multi-scale Adaptive convolutional Pyramid (MACP) for capturing local feature patterns and a Bidirectional Hierarchical Gated Fusion Network (BHGFN) enabling multi-level feature integration across paths, the framework employs a Hierarchical Dynamic Weighting Fusion (HDWF) mechanism to intelligently aggregate multimodal features. Enhanced by a novel Hierarchical Contrastive loss function for feature consistency optimization, UniPTMs demonstrates significant performance improvements (3.2%-11.4% MCC and 4.2%-14.3% AP increases) over state-of-the-art models across five modification types and transcends the Single-Type Prediction Paradigm. To strike a balance between model complexity and performance, we have also developed a lightweight variant named UniPTMs-mini.

# 1.Introduction

As a pivotal mechanism in eukaryotic epigenetic regulation, protein post-translational modifications (PTMs) precisely regulate protein functions through dynamic chemical modifications, participating in cellular signal transduction, metabolic homeostasis, and disease pathogenesis[1,2]. With over 400 identified types spanning conventional modifications (e.g., phosphorylation, glycosylation) and emerging categories (e.g., sulfhydration, carboxyethylation), PTMs orchestrate key biological processes such as cell cycle regulation and stress responses via spatiotemporal-specific modulation of substrate protein activities[3-6]. Aberrant PTM states exhibit strong correlations with major pathologies including tumors and neurodegenerative diseases. Although mass spectrometry and multi-omics technologies have advanced PTM research, conventional approaches remain constrained by temporal resolution limitations, high costs, and insufficient sensitivity for low-abundance modification detection[7,8]. Deep learning-driven predictive models emerge as critical solutions to these challenges, particularly through developing multi-task learning frameworks that integrate synergistic/antagonistic effects across modifications. Such frameworks enable systematic elucidation of PTM dynamic interaction networks, offering novel paradigms for deciphering multi-layered regulatory mechanisms underlying complex diseases.

For post-translational modification (PTM) site prediction, diverse computational strategies have been developed to enhance predictive performance: In N-phosphorylation site prediction, the DeepNphos model integrates One-Hot encoding with EAAC features, achieving high accuracy through a residual convolutional neural network (ResNet) architecture combined with deep transfer learning (DTL)[9]. Comparatively, the Nphos framework employs a GDBT approach by synthesizing amino acid composition, sequence quality indices, and BLOSUM62 matrix features. Detailed descriptions of deep learning models for other PTM types are provided in supplementary materials (Section 1) due to space limitations[10].

A systematic analysis reveals three critical limitations in current deep learning-driven PTM prediction models: First, architectural constraints persist under single-modification-type paradigms, lacking unified representation frameworks to capture cross-type biological feature correlations[11,12]. Second, integration mechanisms between pre-trained embeddings and conventional biological features remain inadequate to overcome modal heterogeneity barriers, resulting in representation conflicts[13,14]. Third, network topologies relying on independent branch architectures exhibit deficiencies in intermediate feature interactions and cross-scale calibration, severely limiting global-local co-optimization capabilities. These compounded limitations ultimately undermine model generalizability and biological interpretability[15-17]. Figure 1 summarizes previous studies on post-translational modification (PTM) site prediction.

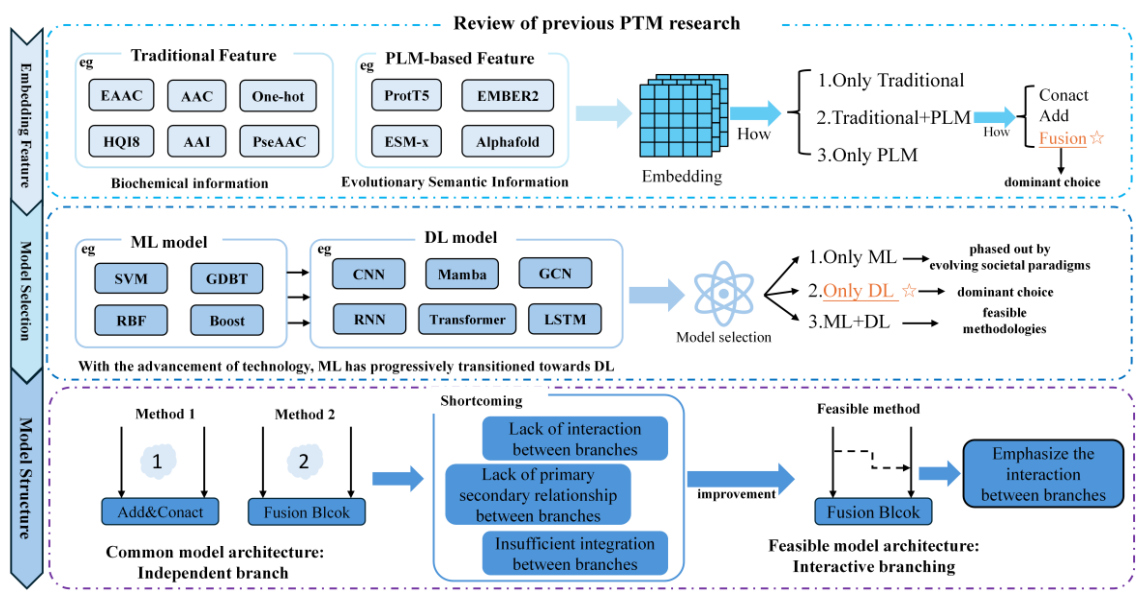

Figure 1. Previous studies on post-translational modification (PTM) site prediction. It presents the embedding techniques, model selection criteria, and architectural frameworks implemented in previous research, while systematically analyzing their methodological limitations and proposing optimization pathways.

To address these challenges, this study innovatively proposes UniPTMs: the first unified multi-PTM type prediction framework, which transcends single modification paradigms through a master-slave collaborative architecture enabling deep multimodal feature integration (see Figure 2 for workflow). The framework implements a three-phase fusion strategy: The master branch employs ProtT5/ESM-2 embeddings as primary inputs, establishing cross-modal interaction pathways via a Bidirectional Gated Cross-Attention (BGCA) module that integrates dynamic gating and multi-scale enhancement for deep semantic fusion. The slave branch combines EMBER2 low-dimensional structural embeddings with conventional features, processed through a Lightweight Low-Dimensional Fusion Network (LDFN) for noise suppression and efficient interaction. A novel Multi-scale Adaptive Convolutional Pyramid (MACP) network is introduced to extract hierarchical sequence features, while the Bidirectional Hierarchical Gated Fusion Network (BHGFN) enables phased intermediate fusion between master-slave branches. This module significantly enhances fusion precision through Asymmetric Hierarchical Attention mechanisms and dynamic convolutional enhancement. The final fusion phase implements a Hierarchical Dynamic Weighting Fusion (HDWF) mechanism, reinforcing master-slave relationships via three-tier weighted integration combined with biologically inspired dynamic regulation. Furthermore, a Hierarchical

Contrastive loss function with intra-layer and cross-layer constraints is designed to strengthen feature consistency. Experimental validation demonstrates the framework's systematic superiority across five PTM site prediction tasks, achieving 3.2%-11.4% MCC and 4.2%-14.3% AP improvements with consistent cross-category performance. UniPTMs represents a critical technological breakthrough for developing universal PTM prediction tools, expanding the application horizons of epigenetic research through its unified analytical capabilities. To strike a balance between model complexity and performance, we have also developed a lightweight variant named UniPTMs-mini.

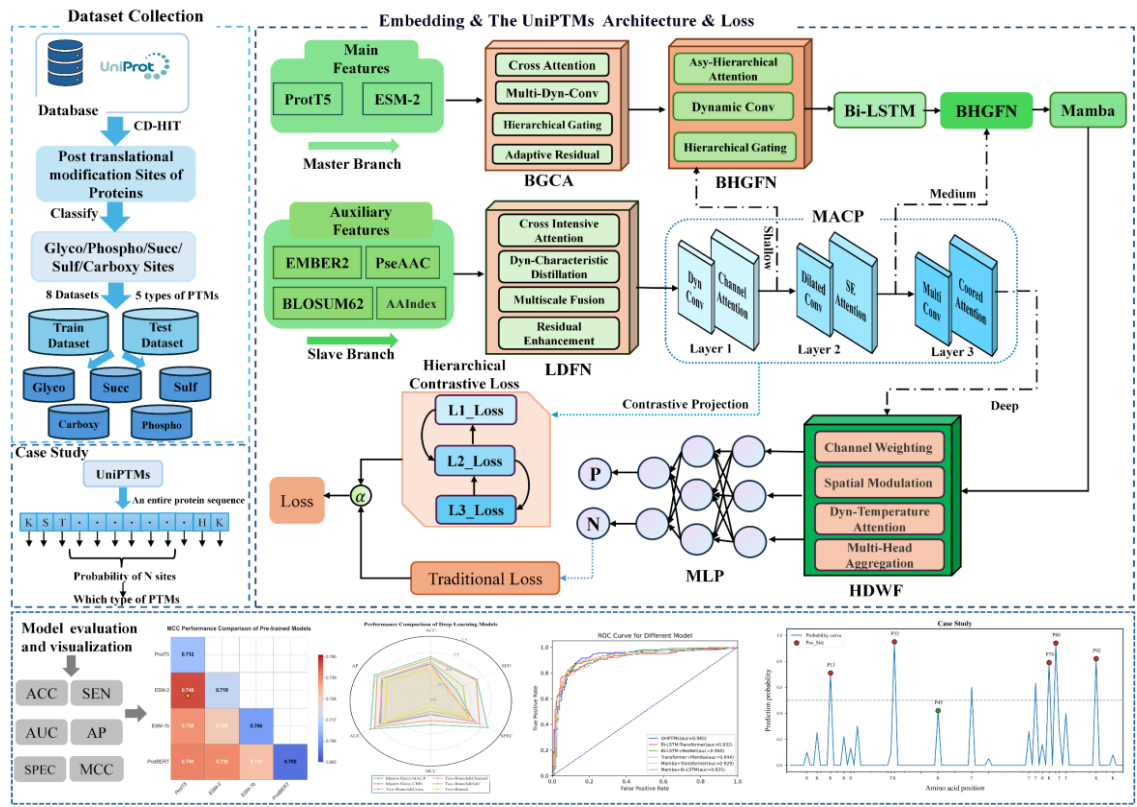

Figure 2. Workflow Diagram of UniPTMs: Step 1. Data Collection; Step 2. Deep Learning Model Construction; Step 3. Model Performance Evaluation; Step 4. Case Study Analysis.

## 2. Material and Methods

### 2.1 Dataset Construction

This study systematically investigated five protein post-translational modification (PTM) types: histidine N-phosphorylation, threonine O-glycosylation, lysine succinylation, cysteine S-carboxyethylation and S-sulfhydration. For the first three well-characterized modifications (N-phosphorylation, O-glycosylation, and succinylation), two independent validation datasets were employed per type to ensure reliability[9-10,18-20][14], while the emerging S-carboxyethylation and

S-sulfhydration modifications were analyzed using single stringently quality-controlled datasets for exploratory investigations[21,22]. All experimental parameters and sequence logo visualizations are provided in Supplementary Materials (Section 2). Detailed dataset statistics including training/test set partitioning and sample size specifications for each PTM type are comprehensively documented in Tables 1-5.

Table 1. Dataset distribution of threonine O-glycosylation (T1/T2: dataset identifiers).

| Datasets | Type | T1 | T2 |
| --- | --- | --- | --- |
| Training set | Positive | 878 | 862 |
|  | Netgative | 878 | 1503 |
| Independent set | Positive | 200 | 216 |
|  | Negative | 200 | 375 |

Table 2. Dataset distribution of histidine N-phosphorylation (H1/H2: dataset identifiers).

| Datasets | Type | H1 | H2 |
| --- | --- | --- | --- |
| Training set | Positive | 512 | 1556 |
|  | Netgative | 512 | 5335 |
| Independent set | Positive | 52 | 389 |
|  | Negative | 52 | 1334 |

Table 3. Dataset distribution of lysine succinylation (K1/K2: dataset identifiers).

| Datasets | Type | K1 | K2 |
| --- | --- | --- | --- |
| Training set | Positive | 10549 | 4755 |
|  | Netgative | 37843 | 50549 |
| Independent set | Positive | 2638 | 254 |
|  | Negative | 9461 | 2977 |

Table 4. Dataset distribution of cysteine S-carboxyethylation (C1: dataset identifier).

| Datasets | Type | C1 |
| --- | --- | --- |
| Training set | Positive | 602 |
|  | Netgative | 602 |
| Independent set | Positive | 358 |
|  | Negative | 3580 |

Table 5. Dataset distribution of cysteine S-sulfhydration (C2: dataset identifier).

| Datasets | Type | C2 |
| --- | --- | --- |
| Training set | Positive | 2164 |
|  | Netgative | 13358 |
| Independent set | Positive | 541 |
|  | Negative | 3339 |

**2.2 Feature Embedding**

This study employed the ProtT5(1024D) and ESM-2(1280D) models to generate protein embeddings as primary features for UniPTMs[23,24], followed by integrating the EMBER2 model (developed by Weissenow et al[25]) with three conventional feature generation methods including

PseAAC, BLOSUM62, and AAIndex to construct auxiliary features[26-28]. Since the initial dimensions of each auxiliary feature are different and such dimensional discrepancies may adversely affect model performance, we performed feature dimension alignment using linear layers and convolutional blocks for the four auxiliary features. Specifically, EMBER2 and PseAAC were projected to a 256-dimensional space, while BLOSUM62 and AAIndex were aligned to a 512-dimensional space. Detailed specifications are provided in Supplementary Materials (Section 3).

**2.3 Model Architecture**

In this study, we adopted a refined modular design strategy for feature fusion processing. The design philosophy of UniPTMs involves equipping both master and slave branches with dedicated fusion modules. The master branch assumes a decisive role, while the slave branch functions as an auxiliary component. These two branches interact through mid-phase fusion modules, culminating in deep integration via post-phase fusion modules. Specifically, for the two primary features (master branch), we achieved comprehensive integration using the BGCA module. Correspondingly, the four auxiliary features (slave branch) underwent deep fusion through the LDFN module. Notably, the BGCA and LDFN modules demonstrate distinct functional orientations: the BGCA module specializes in refined processing of high-dimensional features, whereas the LDFN module focuses on deep exploration of low-dimensional features. Following preliminary fusion of slave branch features, we further implemented the MACP module for multi-level feature extraction, successfully obtaining shallow, intermediate, and deep-level features. The shallow and intermediate features subsequently engage in mid-phase fusion with master branch features through the BHGFN module at predetermined stages. The BHGFN module not only preserves the dominant position of master branch features during fusion but also facilitates efficient collaboration between master and slave features, thereby enhancing overall feature representation. Finally, deep-level features and master branch features processed by the Mamba module undergo post-phase fusion through the HDWF module. By establishing a three-tier weighted mechanism, the HDWF module reinforces inter-branch correlations while empowering the secondary branch to provide supplemental information in critical local regions, thereby achieving more precise and comprehensive feature fusion. The architectural details of UniPTMs are illustrated in Figure 3 and Figure S2.

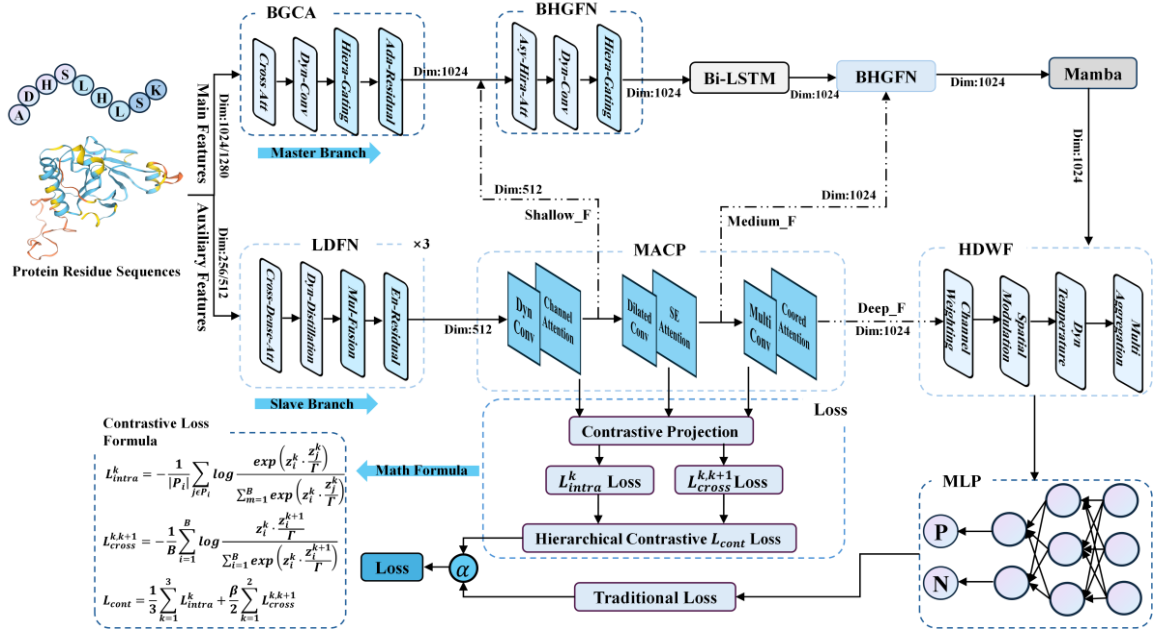

Figure 3. Schematic Diagram of the UniPTMs Framework

**2.3.1 Bidirectional Gated Cross-Attention (BGCA)**

Aiming to address the four critical bottlenecks in protein multimodal fusion methods: information loss, difficulty in capturing dynamic features, fragmented representations and computational complexity, this study proposes a Bidirectional Gated Cross-Attention (BGCA) framework for high-dimensional protein sequence bimodal feature fusion. The innovation establishes cross-modal interactions through bidirectional grouped attention mechanisms, integrating dynamic gating and multi-scale enhancement strategies to achieve deep semantic fusion. Specifically, the framework incorporates multi-scale dynamic convolution and hierarchical gated networks to adaptively capture local-global feature interactions. Additionally, it employs grouped projection techniques combined with residual learning mechanisms to overcome the challenges of high-dimensional fusion complexity.

For input features $X_1$ and $X_2$, cross-modal query ($Q$), key ($K$), and value ($V$) vectors are first generated through group-parameterized projections. Subsequently, bidirectional grouped cross-attention is computed: $Q_1$ derived from ProtT5 interacts with $K_2$ and $V_2$ (from ESM-2) via grouped cross-attention to produce semantically aligned features $Attn_1$. Conversely, through the same mechanism to generate $Attn_2$.

$$Attn_1 = Conact_g \left[ softmax\left(\frac{Q_1(K_2)^T}{\sqrt{d_k}}\right) V_2 \right] \quad (1)$$

$$Attn_2 = Conact_g \left[ softmax\left(\frac{Q_2(K_1)^T}{\sqrt{d_k}}\right) V_1 \right] \quad (2)$$

The bidirectional cross-attention mechanism dynamically facilitates cross-modal integration of evolutionary conservation and local structural accessibility characteristics for PTM sites by leveraging QKV vectors derived from both embeddings(ProtT5 and ESM-2) to enable mutual bidirectional interaction between their feature representations[31,32].

$$C_k = Conv1D(W_k, Attn_1), \quad k \in \{3,5,7\} \tag{3}$$

$$\alpha = softmax(W_g[C_3, C_5, C_7]) \tag{4}$$

$$ConvFusion = \sum_{k \in \{3,5,7\}} \alpha_k \odot C_k \tag{5}$$

The multi-scale dynamic convolution employs 3/5/7-mer convolutional kernels to perform hierarchical extraction of protein features $C_k$. The 3-mer kernel focuses on local physicochemical properties to capture fine grained features, while the 5-mer kernel targets secondary structural motifs within localized regions. The 7-mer kernel models long-range interactions spanning multiple secondary structural units[33]. Furthermore, the dynamic gating mechanism automatically selects biologically significant scales through learnable weights $\alpha$, followed by weighted fusion to generate $ConvFusion$ features[34].

$$\beta = softmax(W_f[ConvFusion; Attn_2]) \tag{6}$$

$$F = \beta_1 \odot ConvFusion + \beta_2 \odot Attn_2 \tag{7}$$

The structural accessibility of PTM sites and their functional conservation exhibit dynamic interdependencies, necessitating context-aware adaptive adjustment of fusion strategies. To achieve fine-grained regulation of feature contributions, we introduce a gating network to generate soft attention weights $\beta$, which dynamically fuse $ConvFusion$ features with $Attn_2$ representations through weighted integration, thereby producing the final fused feature $F$.

**2.3.2 Low-Dimensional Fusion Network (LDFN)**

To address the issues of modal semantic gaps and limitations in shallow interactions inherent in traditional low-dimensional feature fusion, this study proposes a Low-Dimensional Fusion Network (LDFN) for low-dimensional protein feature integration. The framework achieves spatial alignment through an adaptive projection layer to mitigate semantic discrepancies between heterogeneous features. A dual-modal cross-dense attention mechanism is constructed to establish bidirectional guidance channels, enabling dynamic capture of fine-grained correlations. A dynamic gating module is introduced to accomplish sample-adaptive feature distillation. Simultaneously, local patterns extracted by multi-scale convolutional layers are integrated with global features preserved through residual connections. The hierarchical dynamic fusion mechanism ultimately synthesizes these complementary representations through progressive refinement.

For the features $X_1$ and $X_2$ derived from EMBER2 and PseAAC, after aligning their dimensions via linear projection, global interactive features $Attn_i$ are subsequently computed through cross-dense attention. Specifically: $X_1$ serves as the query to retrieve local sequential patterns from $X_2$ ($Q_1 \to KV$), while $X_2$ inversely mines global structural information from $X_1$ ($Q_2 \to KV$). The shared key-value pairs simultaneously reinforce joint representation learning and promote feature coupling.

$$Attn_i = Softmax\left(\frac{Q_i(K)^T}{\sqrt{d}}\right)V, i = 1,2 \qquad (8)$$

Inspired by the refinement principle in knowledge distillation[35], we introduce a dynamic feature distillation mechanism. This approach employs a learnable gating mechanism (with $g$ as the weighting parameter) to adaptively fuse attention-derived features into output D, thereby selectively extracting task-critical information from both modalities while suppressing irrelevant noise.

$$g = \sigma(W_g[Attn_1, Attn_2] + b_g) \qquad (9)$$

$$D = g \odot Attn_1 + (1 - g) \odot Attn_2 \qquad (10)$$

Analogous to the multi-scale dynamic convolutional layers described in the preceding section, this module captures short-, medium-, and long-range dependencies in local protein sequences via distinct kernel sizes ($k$=3,5,7), generating corresponding convolutional features $C_k$. This design compensates for the limitations of attention mechanisms in local feature extraction. Finally, the multi-granularity-enhanced feature $F_c$ is obtained through concatenation followed by a linear transformation to strengthen information integration.

$$C_k = Conv1D(W_k, D), k \in \{3,5,7\} \qquad (11)$$

$$F_c = W_c[C_3, C_5, C_7] + b_c \qquad (12)$$

Finally, the adaptive fusion layer generates the final fused representation through linear combination of the distilled features, multi-scale features, and the additively fused result of the original projected features.

### 2.3.3 Bidirectional Hierarchical Gated Fusion Network (BHGFN)

Given the limitations of traditional mid-level fusion methods in terms of cross-modal interaction depth, feature balance, and topological perception capability, this study innovatively proposes a Bidirectional Hierarchical Gated Fusion Network (BHGFN) for mid-term interaction between master and slave branches. Its core design philosophy establishes a master-slave bidirectional compensation framework that features three key advancements: First, it enhances cross-modal information representation through differentiated asymmetric hierarchical attention mechanisms. Second, it improves topological feature characterization at local modification sites via spatial-channel joint optimized dynamic convolution techniques. Third, it achieves adaptive fusion of multi-granularity features through a novel hierarchical gated network architecture.

For the master branch feature $H_1$ and slave branch feature $H_2$, we establish global dependencies through Asymmetric Hierarchical Attention mechanisms. Specifically, the primary features serve as Query to retrieve semantic information from auxiliary features, generating attention map $A_1$, while conversely, the auxiliary feature adopts the Query role to extract spatial correlation patterns from the primary feature, yielding attention map $A_2$.

$$Attn(Q, K, V) = softmax\left(\frac{Q(K)^T}{\sqrt{d}}\right)V \qquad (13)$$

$$A_1 = Attn(H_1 W_q, K_2, V_2) \qquad (14)$$

$$A_2 = Attn(H_2 W_q, H_1, H_1) \qquad (15)$$

The asymmetric strategy of this attention mechanism is reflected in the attention module $A_1$, which employs $H_2$ projected key-value pairs (KV), while the $A_2$ module retains the original primary features $H_1$. This design ensures that the system consistently maintains $H_1$ as the core reference framework, effectively integrating multi-source information while preserving the stability of primary features.

Subsequently, the attention-level gating mechanism $G_\alpha$ is introduced to dynamically fuse bidirectional features, generating context-aware fused representations $F_\alpha$. Furthermore, a parametric convolutional kernel generator $K_c$ is constructed based on the primary features $H_1$, enabling the convolution operations to acquire local modeling capability adaptive to $H_1$. The dynamic convolution is then applied to $F_\alpha$, enhancing the capability to capture locally abrupt features in sequences, ultimately producing the output $F_c$.

$$G_\alpha = \sigma(W_g[A_1; A_2]) \qquad (16)$$

$$F_\alpha = G_\alpha \odot A_1 + (1 - G_\alpha) \odot A_2 \qquad (17)$$

$$K_c = \phi(AvgPool(H_1)) \qquad (18)$$

$$F_c = Conv1D(F_\alpha, K_c) \qquad (19)$$

The multi-level feature enhancement is achieved through a three-stage gated mechanism: the attention-level gate $F_\alpha$ dynamically regulates bidirectional information interaction ratios, followed by the channel-level gate $F_{gc}$ that filters critical semantic features through channel-wise selection, and finally the spatial-level gate $F_{gs}$ that reinforces sequential positional characteristics through spatial emphasis. These operations are integrated with residual connections to preserve original primary features, collectively generating the enhanced output $Output_f$.

$$G_c = \sigma(f_c(H_1)), F_{gc} = G_c \odot F_c \tag{20}$$

$$G_s = \sigma\left(Conv1d(F_{gc})\right), F_{gs} = G_s \odot F_{gc} \tag{21}$$

$$Output_f = LN(F_{gs} + H_1) \tag{22}$$

**2.3.4 Hierarchical Dynamic Weighting Fusion (HDWF)**

This study proposes a Hierarchical Dynamic Weighting Fusion (HDWF) module for late-stage branch fusion to address the limitations of conventional protein feature fusion methods, such as static weighting coefficients, insufficient information complementarity and training instability. The framework introduces three innovations: a three-level dynamic weighted fusion mechanism (channel-spatial-attentional cascade) guided by a biophysically-inspired temperature regulation strategy; a dual-pathway collaborative modulation approach that combines main-branch channel weighting and auxiliary-branch spatial modulation to achieve cross-hierarchical feature complementarity; and layer-wise scale parameters with post-normalization for training stabilization, further enhanced by a residual dynamic equilibrium mechanism.

For the master branch features M, we employ a channel weighting mechanism (comprised of Global Average Pooling and MLP) to enhance the significance of primary branch features along the channel dimension.

$$g_c = \sigma\left(W_2 \cdot GELU(W_1 \cdot AvgPool(M))\right) \tag{23}$$

$$\tilde{M} = W_p \cdot (M \odot g_c) \tag{24}$$

For the slave branch features S, we implement a spatial weighting mechanism (constructed using components analogous to Spatial Attention) to modulate the spatial dimensions of the secondary branch features. This enhances the spatial specificity of local amino acid environments while capturing localized contextual dependencies.

$$\tilde{S} = S \odot \sigma(W_c \cdot Conv1D(S^T)^T)$$

$\tilde{M}$ and $\tilde{S}$ establish directional feature interaction between master-slave branches through dynamic temperature attention. The biologically inspired dynamic temperature coefficient т is dynamically modulated by the feature standard deviation of the master branch, thereby ensuring the significance of the master branch[36].

$$Q = W_q \tilde{M}, K = W_k \tilde{S} \tag{25}$$

$$\daleth = \alpha + \frac{1}{L}\sum_{i=1}^{L} M_i \tag{26}$$

$$A_{ij} = \frac{\exp\left(\frac{Q_i K_i^T}{\sqrt{D}} \cdot \daleth\right)}{\sum_{k=1}^{L} \exp\left(\frac{Q_i K_i^T}{\sqrt{D}} \cdot \daleth\right)} \tag{27}$$

where $M_i$ denotes the mean feature at the $i$-th position of the main branch, and $A_{ij}$ represents the attention distribution. Subsequently, the output of multi-view feature fusion $\tilde{G}$ is obtained through weighted feature aggregation based on multi-head weight allocation.

$$V = W_v \tilde{M}, G = A \cdot V \tag{28}$$

$$\tilde{G} = \sum_{h=1}^{H} Softmax\left(W_h^{(h)} \cdot GELU(W_m \cdot M)\right) \cdot G^{(h)} \tag{29}$$

where H denotes the number of grouped heads, and A represents the final output of the dynamic temperature attention. Finally, through residual balancing and Layer Scale, the dominance of the main branch is maintained while allowing local refinements:

$$\lambda = \epsilon \cdot I_D \tag{30}$$

$$Out = LN(\lambda \odot M + (1 + \gamma) \cdot \tilde{G} + \beta \cdot M) \tag{31}$$

where $\gamma$ and $\beta$ are learnable parameters that dynamically adjust the blending ratio between fused features and the main branch. $\lambda$ denotes the layer scale factor to enhance stability, $\epsilon$ is initialized to 1e-6, and $I_D$ represents the scaling factor[37].

**2.3.5 Multi-Scale Adaptive Convolution Pyramid (MACP)**

Given the multi-level cooperative regulation of protein functionality through local motifs, domains, and global folding hierarchies, this study proposes a Multi-scale Adaptive Convolution Pyramid (MACP), which enables cross-hierarchy mapping via three-stage hierarchical feature learning. The shallow stage integrates dynamic convolutional kernels with Channel Attention mechanisms to extract localized physicochemical patterns. The intermediate stage combines dilated residual convolutions for modeling long-range dependencies of functional modules and SE Attention for discriminative region localization. The deep stage synergizes multi-scale convolutions with

Coordinate Attention to jointly characterize residue-specific channel-position interdependencies. A cross-scale adaptive fusion mechanism dynamically aggregates multi-granularity features through learnable hierarchical weighting.

For the output feature X processed by the LDFN module, it is first passed through the shallow branch where dynamic convolution adaptively selects convolution kernels of length 3 or 5 to capture local sequential patterns across varying ranges (with $\alpha$ denoting the weighting factor). A channel attention mechanism is simultaneously employed to enhance responses in critical channels:

$$\alpha = \sigma(W_s \cdot GAP(X)) \tag{32}$$

$$F_1 = Channel(\alpha \cdot Conv3(X) + (1-\alpha) \cdot Conv5(X)) \tag{33}$$

Subsequently, the feature $F_1$ is concatenated and propagated to the middle branch, where dilated residual convolution ($Conv_{dil}$) is employed to expand the receptive field for contextual information capture. This process is integrated with an SE attention mechanism to emphasize salient regions[38], ultimately generating feature $F_2$:

$$F_{main} = BN\left(GELU(Conv_{dil}(F_1))\right) \tag{34}$$

$$F_2 = SE(F_{main}) + Conv(F_1) \tag{35}$$

Feature $F_2$ is subsequently fed into the deep branch, where multi-scale convolutional layers (with kernel sizes {3,5,7}) perform hierarchical feature extraction to obtain output $F_c$. A Coordinate Attention mechanism is then introduced to seamlessly integrate spatial coordinate information into channel-wise attention[39], enabling simultaneous enhancement of cross-channel and spatial correlations. Here, $A_c$ denotes channel attention weights while $A_s$ represents spatial attention weights:

$$A_c = \sigma\left(W_2 \cdot \delta(W_1 \cdot GAP(F_c))\right), A_s = \sigma(Conv_{3\times3}(F_c)) \tag{36}$$

$$F_3 = F_2 \odot (A_c \otimes A_s)$$

Where $F_3$ denotes the output of the deep branch. $F_1$, $F_2$, and $F_3$ are progressively integrated with the main branch across different stages. Furthermore, these three outputs are employed to compute the hierarchical contrastive loss.

**2.3.6 Bi-LSTM and Mamba**

This study employs a sequential integration strategy combining Bidirectional Long Short-Term Memory (Bi-LSTM) and Mamba for protein feature extraction[40,41]. First, Bi-LSTM's bidirectional architecture captures shallow features, utilizing gating mechanisms to suppress local noise while emphasizing critical sites. Subsequently, Mamba which is a structured state space model, extracts mid-level features through implicit state transitions and selective scanning, dynamically capturing long-range dependencies with linear computational complexity that prevents gradient vanishing, making it particularly effective for modeling cross-regional sequence patterns. Detailed descriptions of Bi-LSTM and Mamba architectures are provided in the Supplementary Materials (Section 4).

**2.4 Loss Function**

Inspired by contrastive learning principles, this study innovatively proposes a Hierarchical Contrastive Loss function that establishes a multi-granularity feature interaction framework through systematic integration of intra-layer feature contrast and cross-layer semantic contrast mechanisms[42,43]. The proposed function incorporates dynamic temperature modulation and feature normalization strategies to stabilize the training process, while implementing cross-hierarchical positive pair construction and interference sample filtering mechanisms. These components synergistically enhance intra-class representation consistency while ensuring semantic coherence across multi-level feature spaces. The mathematical formulation of this hierarchical contrastive loss function is presented below:

$$L_{intra}^k = -\frac{1}{|P_i|} \sum_{j \in P_i} \log \frac{\exp\left(z_i^k \cdot \frac{z_j^k}{\Gamma}\right)}{\sum_{m=1}^{B} \exp\left(z_i^k \cdot \frac{z_j^k}{\Gamma}\right)} \tag{37}$$

Here, $L_{intra}^k$ denotes the intra-layer contrastive loss, where $P_i$ represents the positive sample set, $\Gamma$ is a learnable temperature parameter to enhance the compactness of the feature space, $k \in \{1,2,3\}$ corresponds to shallow, intermediate, and deep-layer features, respectively, and $B$ denotes the batch size. The core principle of the intra-layer loss function is to minimize the distance between samples of the same class while maximizing the separation between dissimilar samples within the feature space of the same network layer. This is followed by the cross-layer contrastive loss:

$$L_{cross}^{k,k+1} = -\frac{1}{B} \sum_{i=1}^{B} \log \frac{z_i^k \cdot \frac{z_i^{k+1}}{\Gamma}}{\sum_{i=1}^{B} \exp\left(z_i^k \cdot \frac{z_i^{k+1}}{\Gamma}\right)} \tag{38}$$

Here, $L_{cross}^{k,k+1}$ denotes the cross-layer contrastive loss. The core principle of this loss is to minimize the distance between multi-level features of the same sample while maximizing the separation between features of different samples across hierarchical feature spaces, thereby establishing hierarchical feature consistency constraints. The total contrastive loss $L_{cont}$ is subsequently obtained through a weighted combination of intra-layer and cross-layer contrastive losses:

$$L_{cont} = \frac{1}{3}\sum_{k=1}^{3} L_{intra}^{k} + \frac{\beta}{2}\sum_{k=1}^{2} L_{cross}^{k,k+1} \tag{39}$$

Here, $\beta$ denotes a tunable weighting parameter, with experimental results demonstrating optimal performance when $\beta = 0.7$. For the three outputs $F_{1,2,3}$ of the MACP module, we compute the contrastive loss $L_{cont}$, while the conventional loss $L_c$ is derived from the main branch output $F$. The total loss function is formulated as:

$$L_{total} = L_c + \lambda(t)L_{cont} \tag{40}$$

Where $\lambda(t)$ represents a dynamic weighting coefficient that enables progressive decoupling of primary and auxiliary tasks during training.

**2.5 Hyperparameter Setting and Model Evaluation**

To evaluate the performance of our deep learning model, the following metrics were adopted: accuracy (ACC), sensitivity (SEN), specificity (SPEC), the Matthews correlation coefficient (MCC), and average precision (AP). The mathematical formulations of these metrics along with the hyperparameter configurations of the model are detailed in the Supplementary Materials (Section 5-6).

## 3.Results

In this study, we conducted an ablation analysis focusing on threonine O-glycosylation site prediction using the T2 dataset, followed by a systematic comparison of UniPTMs with five categories of post-translational modification (PTM) prediction models across eight benchmark datasets to comprehensively evaluate their predictive capabilities.

### 3.1 Ablation Experiment

#### 3.1.1 Evaluation of Contributions from Pre-trained Model Embeddings to Predictive Performance

This study systematically evaluated four protein pre-trained models (ProtT5, ESM-2, ESM-1b, and ProtBERT) through orthogonal experimental design and cross-validation, specifically analyzing the complementary effects of pairwise combinations of their sequence embedding features and assessing their applicability within the UniPTMs framework. Based on MCC value ranking and heatmap visualization analysis (Table 6, Figure 4), the combination of ProtT5 and ESM-2 demonstrated optimal performance with an MCC value of 0.746. Furthermore, we propose a plug and play auxiliary feature module that incorporates EMBER2, PseAAC, BLOSUM62, and AAIndex as optional complementary features to enhance framework flexibility. If researchers need to replace features with those more suitable for their specific research, they can do so conveniently.

Table 6. MCC values of embedding combinations generated by different protein pre-trained models

| Model | ProtT5 | ESM-2 | ESM-1b | ProtBERT |
|---|---|---|---|---|
| **ProtT5** | 0.712 | | | |
| **ESM-2** | **0.746** | 0.718 | | |
| **ESM-1b** | 0.738 | 0.729 | 0.706 | |
| **ProtBERT** | 0.740 | 0.739 | 0.731 | 0.700 |

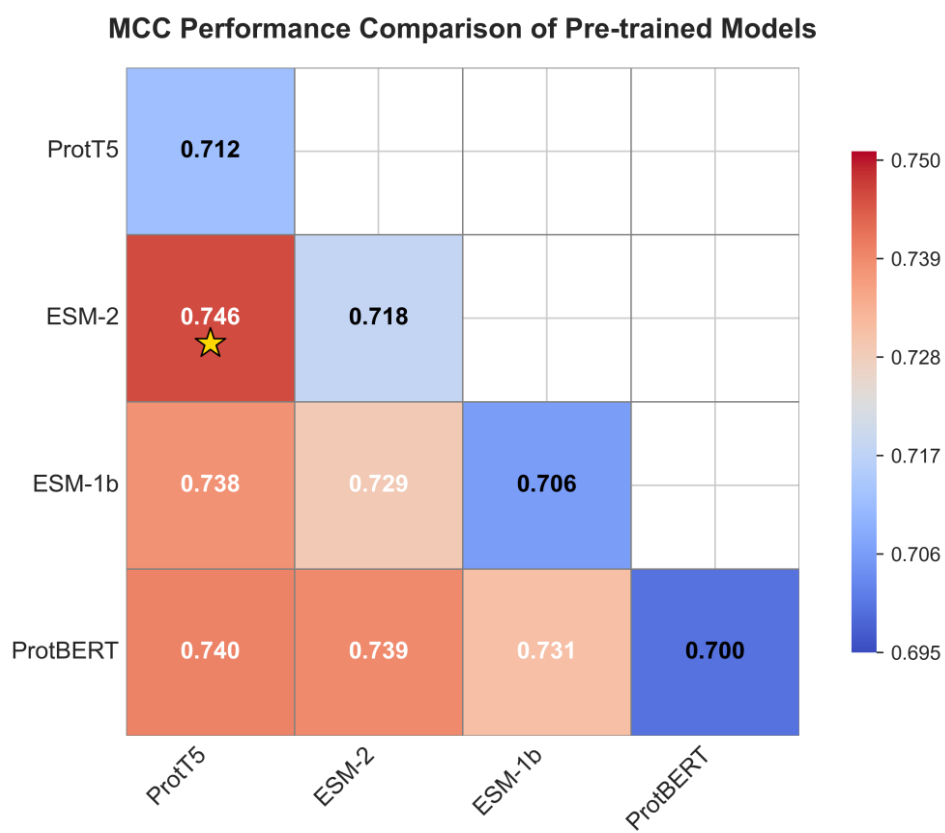

Figure 4. Heatmap of MCC Values for Embedding Combinations Generated by Protein Pre-trained Models.

**3.1.2 Contribution Analysis of Master-Slave Architecture to Framework Performance**

This study substantiated the technical superiority of the master-slave architecture and its specialized MACP module through systematic multi-dimensional comparative experiments. Employing a dual-branch architecture as the baseline reference, we constructed four distinct experimental configurations by integrating Cross-Attention, Channel-Attention, Self-Attention, and basic fusion methodologies. Rigorous quantitative analyses conducted via controlled variable methodologies, detailed in Supplementary Table 7 and Figure 5, yielded two critical insights: First, the proposed master-slave framework demonstrated marked performance enhancements over cross-attention based fusion modules, registering absolute improvements of 6.2 percentage points in Matthews Correlation Coefficient (MCC), 5% in Area Under the Curve (AUC), and 4.6% in Average Precision (AP). Second, the novel MACP module exhibited hierarchical feature abstraction capabilities that surpassed conventional CNN architectures, achieving measurable gains of 2.6% (MCC), 2.5% (AUC), and 1.5% (AP) through its structured processing mechanism.

Table 7. Comparative Analysis of Master-Slave Architecture versus Mainstream Conventional Architectures

| Model | ACC | SEN | SPEC | MCC | AUC | AP |
| --- | --- | --- | --- | --- | --- | --- |
| **Master-Slave-MACP** | **0.873** | **0.865** | **0.925** | **0.746** | **0.945** | **0.921** |
| **Master-Slave-CNN** | 0.860 | 0.860 | 0.880 | 0.720 | 0.920 | 0.906 |
| **Two-Branch&Cross** | 0.833 | 0.745 | 0.845 | 0.684 | 0.895 | 0.875 |
| **Two-Branch&Channel** | 0.828 | 0.710 | 0.855 | 0.680 | 0.889 | 0.882 |
| **Two-Branch&Self** | 0.845 | 0.785 | 0.810 | 0.662 | 0.885 | 0.878 |
| **Two-Branch&Conact** | 0.800 | 0.720 | 0.825 | 0.637 | 0.862 | 0.850 |

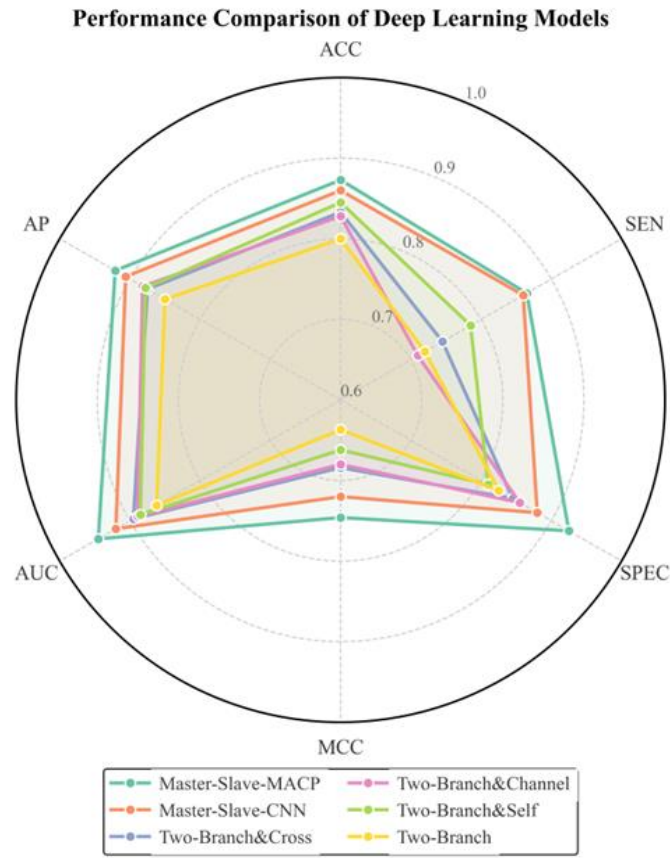

Figure 5. Radar Chart of Performance Comparison Across Architectural Variants

**3.1.3 Ablation Analysis of Stage-Wise Fusion Modules**

The UniPTMs framework incorporates a three-stage fusion mechanism, consisting of early-stage (BGCA + LDFN), middle-stage (BHGFN), and late-stage (HDWF) modules. To systematically evaluate each component's contribution, we conducted ablation studies by progressively replacing these stage-specific modules with conventional additive or concatenation operations. Quantitative comparisons through tabular metrics (Table 8) and bar chart visualization (Figure 6) revealed progressive performance enhancements: Early-stage fusion improved MCC/AUC/AP by 3.1%/2.3%/2.0%; Middle-stage integration boosted gains to 5.4%/3.1%/2.5%; Late-stage refinement further increased improvements to 5.6%/3.5%/3.1%. The fully integrated framework achieved state-of-the-art performance with MCC = 0.746, AUC = 0.945, and AP = 0.921. These results demonstrate that hierarchical multi-stage fusion effectively consolidates complementary protein representations across temporal scales.

Table 8. Ablation Study Results of Multi-stage Fusion Modules

| Model | ACC | SEN | SPEC | MCC | AUC | AP |
| --- | --- | --- | --- | --- | --- | --- |
| **UniPTMs** | **0.873** | **0.865** | 0.925 | **0.746** | **0.945** | **0.921** |
| **None Early Interation** | 0.865 | 0.835 | 0.885 | 0.715 | 0.922 | 0.901 |
| **None Middle Interation** | 0.858 | 0.845 | 0.870 | 0.692 | 0.914 | 0.896 |
| **None Late Interation** | 0.840 | 0.835 | 0.870 | 0.690 | 0.910 | 0.890 |
| **Only Early Interation** | 0.810 | 0.714 | 0.816 | 0.620 | 0.885 | 0.872 |
| **Only Middle Interation** | 0.833 | 0.810 | 0.810 | 0.648 | 0.898 | 0.880 |
| **Only Late Interation** | 0.830 | 0.720 | 0.860 | 0.642 | 0.893 | 0.875 |
| **Add or Conact** | 0.809 | 0.655 | **0.945** | 0.618 | 0.875 | 0.861 |

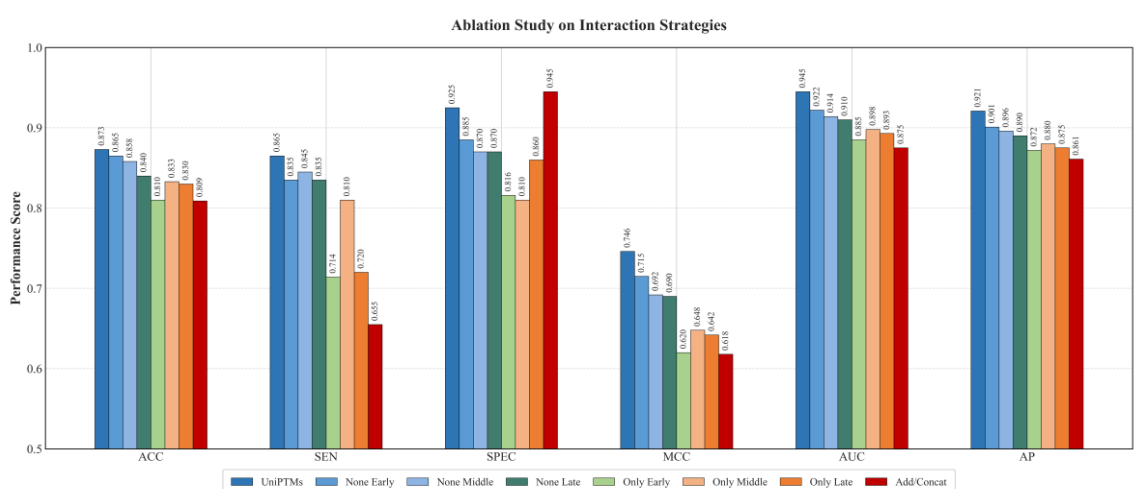

Figure 6. Comparative Bar Chart of Ablation Study for Multi-stage Fusion Modules

### 3.1.4 Selection of Deep Learning Model

This study conducts quantitative evaluation and optimization of a dual-model cascaded architecture for protein feature extraction through a systematic ablation experimental framework. The research establishes a benchmark system integrating four deep learning models: Bi-LSTM, Mamba, Transformer, and ResNet, with particular focus on investigating the synergistic effects of heterogeneous model combinations on feature extraction enhancement, as well as the influence of cascade sequence on temporal dependencies in representation learning. Experimental results demonstrate that the proposed Bi-LSTM-Mamba collaborative framework (UniPTMs) exhibits significant advantages in protein feature representation learning, achieving superior performance metrics including Matthew's correlation coefficient (MCC=0.746), area under ROC curve (AUC=0.945), and average precision (AP=0.921), consistently outperforming other model combinations. The hierarchical feature extraction mechanism of this framework demonstrates biological plausibility: Bi-LSTM captures local structural patterns in protein sequences through gated memory units, while Mamba models long-range dependencies across structural domains using dynamic weight mechanisms in state space models, synergistically enabling efficient transition from primary feature extraction to intermediate feature fusion (detailed experimental data presented in Table 9 and ROC/PR curve analysis in Figure 7).

Table 9. Comparison Results of Deep Learning Models

| Model | ACC | SEN | SPEC | MCC | AUC | AP |
| --- | --- | --- | --- | --- | --- | --- |
| **UniPTMs** | **0.873** | **0.865** | **0.925** | **0.746** | **0.945** | **0.921** |
| **Bi-LSTM-Transformer** | 0.868 | 0.845 | 0.895 | 0.736 | 0.932 | 0.913 |
| **Bi-LSTM+ResNet** | 0.865 | 0.840 | 0.895 | 0.730 | 0.940 | 0.900 |
| **Transformer+Mamba** | 0.845 | 0.830 | 0.895 | 0.728 | 0.944 | 0.893 |
| **Mamba+Transformer** | 0.830 | 0.825 | 0.880 | 0.711 | 0.929 | 0.886 |
| **Mamba+Bi-LSTM** | 0.815 | 0.830 | 0.876 | 0.706 | 0.925 | 0.880 |

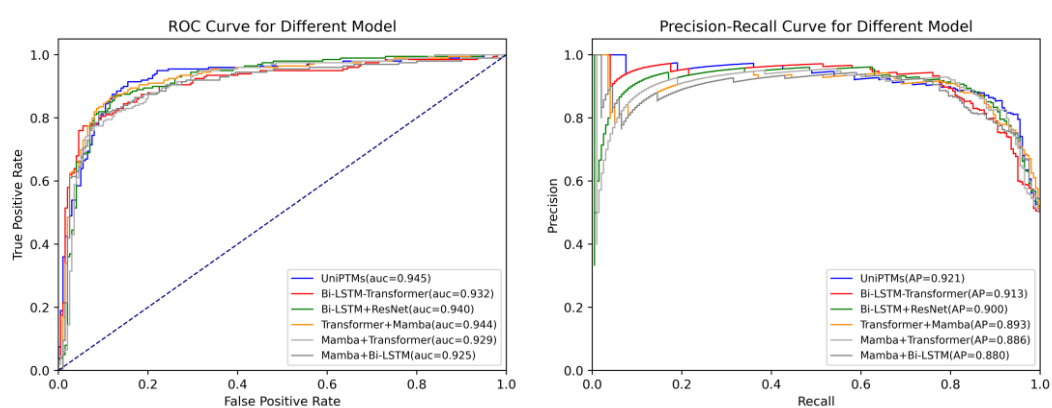

Figure 7. ROC and PR curves of different model combinations

**3.1.5 Evaluate the effectiveness of Hierarchical Contrast Loss function**

To evaluate the effectiveness of the proposed Hierarchical Contrastive Loss (HCL), this study designs systematic comparative experiments tailored to the class imbalance characteristics of the O-linked threonine glycosylation site dataset. Within the traditional supervised learning framework, Focal Loss (FL) and Weighted Cross-Entropy Loss (WCE) are adopted as baselines, while the contrastive learning framework incorporates the self-designed HCL and InfoNCE loss for cross-paradigm comparisons (experimental details are provided in Table 10 and Figure 8). Results demonstrate that the HCL+FL combination achieves optimal performance on key metrics, with the Matthews Correlation Coefficient (MCC) reaching 0.746 (a 1.6% improvement) and Average Precision (AP) attaining 0.921 (a 0.5% gain). Although the HCL+WCE pairing retains a marginal 0.3% advantage in Area Under the Curve (AUC), AP is prioritized as it directly reflects positive class prediction efficacy. The HCL+FL synergy is ultimately selected as the optimal configuration, with all hybrid training paradigms (HCL+FL/WCE) exhibiting systematic superiority: MCC improves by 2.4–3.5%, and AP increases over 2%, empirically validating the theoretical benefits of contrastive learning mechanisms in feature disentanglement and model enhancement.

Table 10. Comparative Performance Evaluation of Loss Functions and Their Combinations

|  | ACC | SEN | SPEC | MCC | AUC | AP |
|---|---|---|---|---|---|---|
| **HCL+FL(UniPTMs)** | **0.873** | 0.865 | 0.925 | **0.746** | 0.945 | **0.921** |
| **HCL+WCE** | 0.870 | **0.875** | 0.910 | 0.730 | **0.948** | 0.916 |
| **InfoNCE+FL** | 0.853 | 0.855 | 0.930 | 0.720 | 0.924 | 0.910 |
| **InfoNCE+WCE** | 0.863 | 0.840 | 0.925 | 0.715 | 0.912 | 0.909 |
| **Focal Loss** | 0.845 | 0.835 | **0.940** | 0.711 | 0.923 | 0.901 |
| **WCE Loss** | 0.833 | 0.850 | 0.905 | 0.706 | 0.909 | 0.896 |

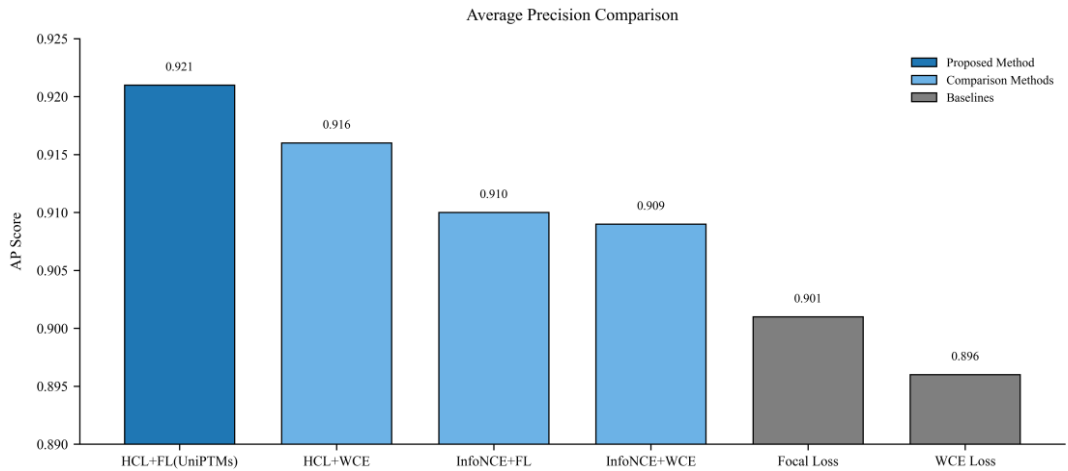

Figure 8. Comparative analysis of AP values using bar charts for different loss functions and their combinations.

### 3.2 Visualization of Feature Extraction Process

The UMAP three-dimensional visualization (Figure 9) revealed that positive and negative samples were intermixed and indistinguishable in the original dataset, whereas distinct boundaries between the samples emerged after processing through the primary-secondary architecture of UniPTMs. Moreover, visualization validation on the test set demonstrated that the model exhibited superior performance across both training and testing datasets, indicating that the features extracted by UniPTMs possess significant effectiveness in identifying O-glycosylated threonine sites.

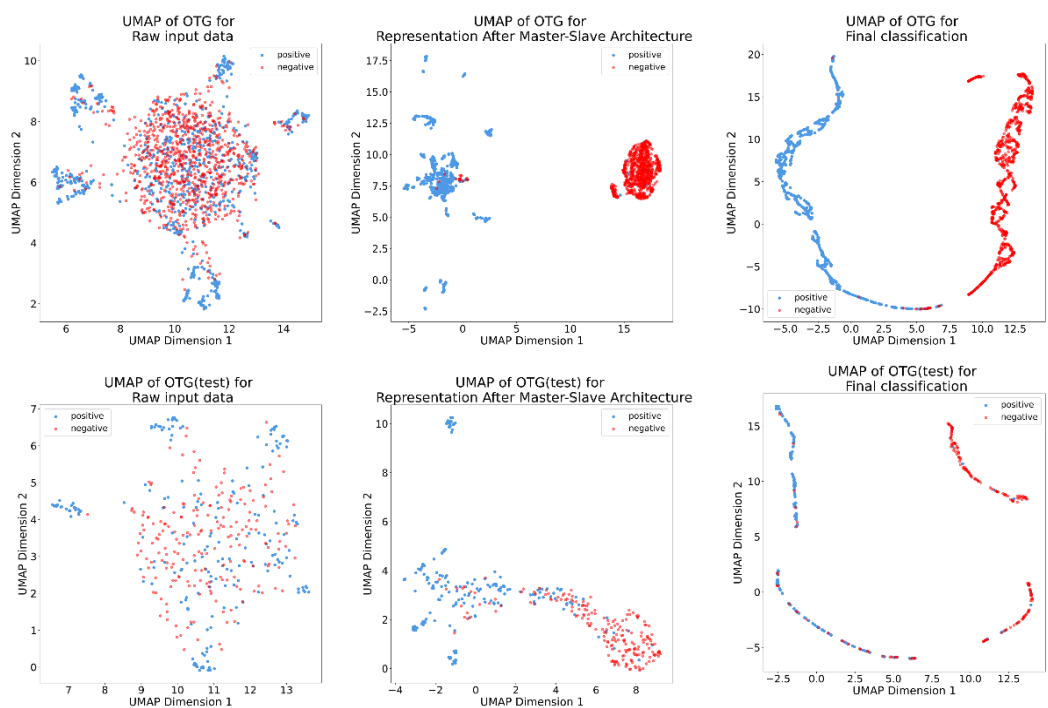

Figure 9. Visualization process of feature extraction of threonine O-glycosylation sites based on UMAP. The upper part is based on the train dataset, and the lower part is based on the test dataset.

## 3.3 Independent testing of UniPTMs for Unified Multi-PTM type prediction framework

### 3.3.1 Comparison between UniPTMs and existing prediction models of various PTM types

This study breaks through the conventional single-modification-type prediction paradigm by establishing a multi-dimensional validation framework covering five critical types of post-translational modifications (PTMs). Utilizing eight independent datasets: spanning O-glycosylation, N-phosphorylation, Succinylation, and other modifications, and encompassing diverse sample characteristics such as balanced/imbalanced distributions and varying sample sizes, the UniPTMs model was systematically validated for its generalizability and efficacy in predicting multi-type PTM sites. At the model construction level, open-source toolkits were selectively curated for each modification type based on accessibility principles: Threonine O-glycosylation integrated five models [19,44-47],Histidine N-phosphorylation incorporated four models[9-10,48-49],Lysine succinylation analyzed six models [20,50-54],Cysteine S-carboxyethylation employed the sole available model DLBWE-Cys[21],S-sulfhydration sites combined two models[22,55].All non-public or access restricted tools were excluded. Detailed performance comparisons are provided in Tables 10-14.

Table 10. Performance comparison of various models for predicting Threonine O-Glycosylation sites on T1/T2 Dataset

| Evaluation on T1 Dataset(balanced) | | | | | | |
|---|---|---|---|---|---|---|
| **Model** | **ACC** | **SEN** | **SPEC** | **MCC** | **AUC** | **AP** |
| **UniPTMs** | 0.871 | **0.880** | 0.905 | **0.785** | **0.940** | 0.914 |
| **DOGpred** | 0.874 | 0.845 | **0.915** | 0.742 | 0.922 | 0.820 |
| **HOTGpred** | **0.882** | 0.840 | 0.902 | 0.738 | 0.920 | 0.815 |
| **NetOGlyc-4.0** | 0.800 | 0.790 | 0.855 | 0.661 | 0.846 | 0.603 |
| **GlycoMine** | 0.520 | 0.660 | 0.415 | 0.102 | 0.688 | 0.222 |
| **OGP** | 0.761 | 0.822 | 0.693 | 0.555 | 0.812 | 0.573 |
| Evaluation on T2 Dataset(imbalanced) | | | | | | |
| **UniPTMs** | **0.873** | 0.865 | **0.925** | **0.746** | **0.945** | **0.921** |
| **DOGpred** | 0.866 | 0.830 | 0.908 | 0.640 | 0.919 | 0.818 |
| **HOTGpred** | 0.861 | 0.842 | 0.887 | 0.632 | 0.916 | 0.811 |
| **NetOGlyc-4.0** | 0.400 | 0.772 | 0.844 | 0.515 | 0.861 | 0.572 |
| **GlycoMine** | 0.192 | **0.995** | 0.034 | 0.075 | 0.646 | 0.168 |
| **OGP** | 0.702 | 0.833 | 0.690 | 0.389 | 0.802 | 0.500 |

Table 11. Performance comparison of various models for predicting Histidine N-Phosphorylation sites on H1/H2 Dataset

| Evaluation on H1 Dataset(balanced) | | | | | | |
|---|---|---|---|---|---|---|
| **Model** | **ACC** | **SEN** | **SPEC** | **MCC** | **AUC** | **AP** |
| **UniPTMs** | **0.845** | 0.842 | **0.865** | **0.705** | **0.919** | **0.890** |
| **Nphos** | 0.828 | **0.856** | 0.769 | 0.630 | 0.876 | 0.843 |
| **GPS-pPLM** | 0.803 | 0.810 | 0.810 | 0.606 | 0.848 | 0.821 |
| **DeepNphos** | 0.717 | 0.761 | 0.673 | 0.473 | 0.802 | 0.726 |
| **pHisPred** | 0.644 | 0.692 | 0.596 | 0.290 | 0.599 | 0.647 |
| Evaluation on H2 Dataset(imbalanced) | | | | | | |
| **UniPTMs** | **0.878** | **0.810** | 0.875 | **0.734** | **0.910** | **0.909** |
| **Nphos** | 0.876 | 0.630 | **0.952** | 0.655 | 0.891 | 0.863 |
| **GPS-pPLM** | 0.820 | 0.762 | 0.831 | 0.626 | 0.860 | 0.831 |
| **DeepNphos** | 0.759 | 0.693 | 0.762 | 0.510 | 0.835 | 0.806 |
| **pHisPred** | 0.696 | 0.597 | 0.721 | 0.410 | 0.681 | 0.667 |

Table 12. Performance comparison of various models for predicting Lysine Succinylation sites on K1/K2 Dataset

| | Evaluation on K1 Dataset | | | | | |
|---|---|---|---|---|---|---|
| **Model** | **ACC** | **SEN** | **SPEC** | **MCC** | **AUC** | **AP** |
| **UniPTMs** | 0.793 | **0.680** | 0.867 | **0.480** | 0.796 | **0.707** |
| **pSuc-EDBAM** | 0.789 | 0.369 | 0.919 | 0.338 | 0.806 | 0.518 |
| **LMSuccSite** | **0.813** | 0.342 | 0.944 | 0.366 | **0.825** | 0.564 |
| **Deep_KsuccSite** | 0.754 | 0.489 | 0.828 | 0.306 | 0.767 | 0.452 |
| **Deep-Ksucc** | 0.806 | 0.315 | 0.942 | 0.335 | 0.815 | 0.527 |
| **LSTMCNNsucc** | 0.805 | 0.288 | **0.949** | 0.323 | 0.822 | 0.543 |
| **DeepSuccinylSite** | 0.774 | 0.185 | 0.942 | 0.189 | 0.734 | 0.406 |
| | Evaluation on K2 Dataset | | | | | |
| **UniPTMs** | 0.769 | 0.735 | 0.730 | **0.442** | 0.805 | **0.655** |
| **pSuc-EDBAM** | 0.743 | 0.710 | 0.722 | 0.292 | **0.819** | 0.497 |
| **LMSuccSite** | 0.783 | **0.745** | 0.700 | 0.328 | 0.815 | 0.535 |
| **Deep_KsuccSite** | 0.705 | 0.553 | 0.707 | 0.262 | 0.722 | 0.442 |
| **Deep-Ksucc** | 0.786 | 0.690 | 0.746 | 0.346 | 0.801 | 0.488 |
| **LSTMCNNsucc** | **0.799** | 0.348 | **0.868** | 0.305 | 0.806 | 0.549 |
| **DeepSuccinylSite** | 0.745 | 0.375 | 0.852 | 0.200 | 0.697 | 0.410 |

Table 13. Performance Comparison of Models for Predicting Cysteine S-Carboxyethylation Sites

| **Model** | **ACC** | **SEN** | **SPEC** | **MCC** | **AUC** | **AP** |
|---|---|---|---|---|---|---|
| **UniPTMs** | **0.775** | **0.723** | 0.843 | **0.573** | **0.816** | **0.858** |
| **DLBWE-Cys** | 0.753 | 0.656 | **0.854** | 0.515 | 0.790 | 0.816 |

Table 14. Performance Comparison of Models for Predicting Cysteine S-Sulfhydration Sites

| **Model** | **ACC** | **SEN** | **SPEC** | **MCC** | **AUC** | **AP** |
|---|---|---|---|---|---|---|
| **UniPTMs** | 0.752 | 0.730 | 0.768 | **0.572** | **0.789** | **0.842** |
| **Sul-BertGRU** | **0.770** | **0.848** | 0.680 | 0.540 | 0.760 | 0.752 |
| **pCysMod** | 0.758 | 0.715 | **0.775** | 0.510 | 0.778 | 0.790 |

The analytical results demonstrate that UniPTMs achieves comprehensive and superior performance in predicting diverse protein modification sites: For O-glycosylated threonine sites on both balanced and imbalanced datasets, UniPTMs improved the MCC, AUC, and AP metrics by 4.3%–4.6%, 1.8%–2.6%, and 9.4%–10.3%, respectively, compared to state-of-the-art models. For N-phosphorylated histidine sites, the corresponding improvements reached 7.5%–7.9% (MCC), 1.9%–4.3% (AUC), and 4.6%–4.7% (AP). For lysine succinylation sites, MCC and AP were enhanced by 9.6%–11.4% and 10.6%–14.3%, respectively. For cysteine S-carboxyethylation and S-sulfhydration sites, the improvements in MCC, AUC, and AP ranged from 3.2%–5.8%, 1.1%–2.7%, and 4.2%–5.2%, respectively. These data robustly validate the significant advantage of UniPTMs in predicting multiple types of protein modification sites. Furthermore, visualizations via dot plots and stacked bar charts are provided (Figure 10 and Figure 11).

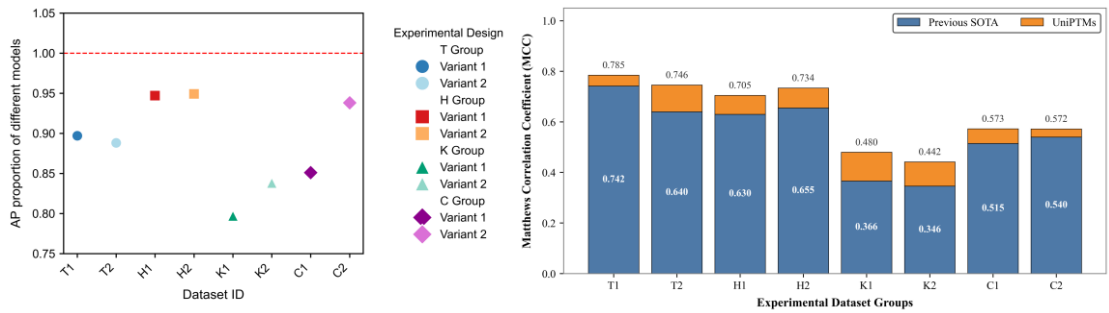

Figure 10. Dot plot and stacked bar chart. Labels on the X-axis represent dataset IDs, while the Y-axis indicates the AP ratio (calculated as the AP value of the state-of-the-art model for each PTM type divided by that of UniPTMs) and MCC values.

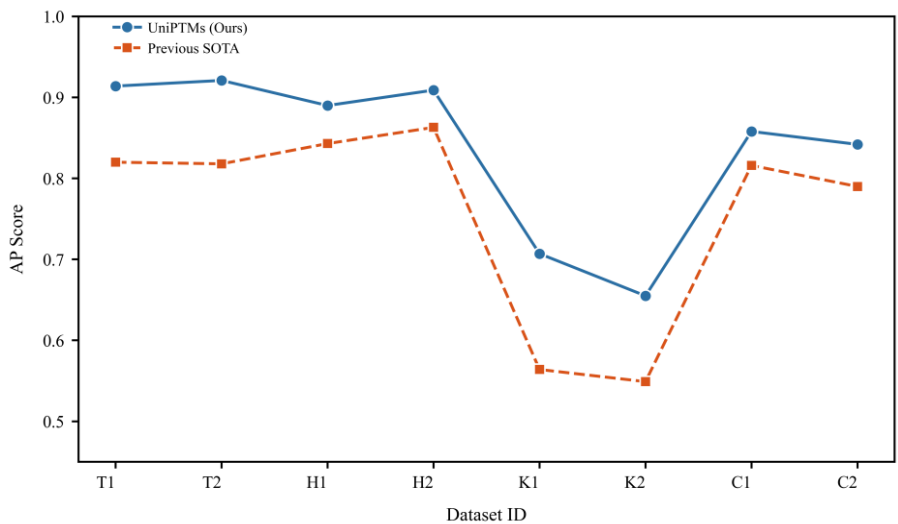

Figure 11. Line chart. The x-axis represents the dataset indices, and the y-axis denotes the AP values. As can be seen from the figure, UniPTMs outperforms previous SOTA models across all datasets.

To rigorously evaluate model performance while mitigating interference from hyperparameters and other uncontrollable variables, we conducted two rounds of 5-fold cross-validation across 3 categories of prevalent PTMs (encompassing 6 datasets), employing MCC (Matthews Correlation Coefficient) as the primary evaluation metric. Visual analysis of MCC scatter plots (Figure 12) demonstrated that all data points corresponding to UniPTMs models were consistently distributed above the y=x baseline, experimentally confirming that UniPTMs achieved significantly higher MCC values than comparative models while maintaining stability at elevated performance levels. This breakthrough transcends conventional single modification type prediction paradigms, providing an innovative bioinformatics toolkit for systematic multi-dimensional PTMs research.

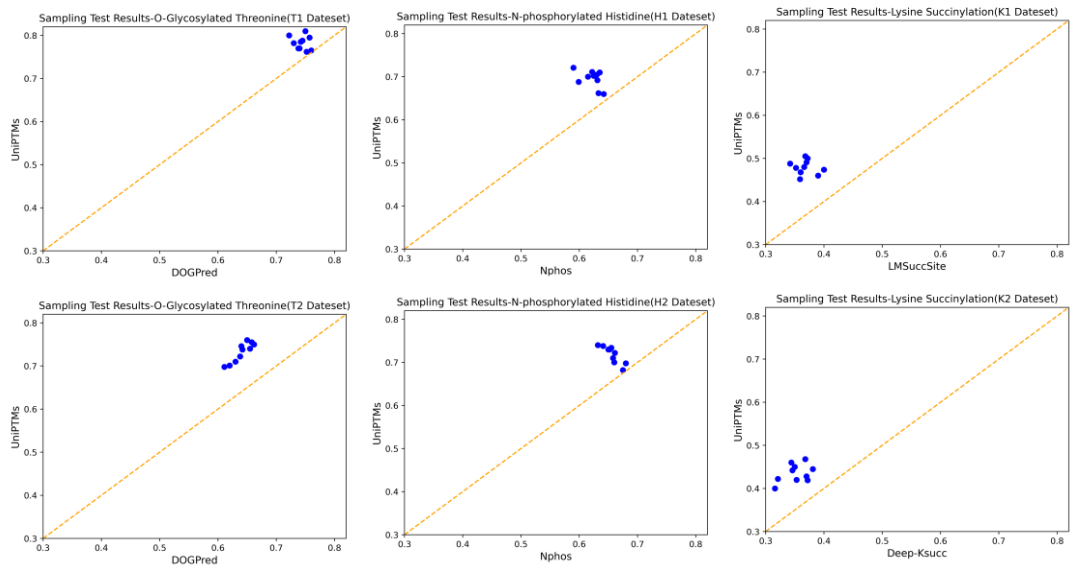

Figure 12. comparison of MCC indicators of the performance of existing deep learning models under each data set, X coordinate is the MCC of x-axis model, y coordinate is the MCC of y-axis model.

### 3.3.2 Case Study

This study addresses the limitations of conventional post-translational modification (PTM) prediction methods that focus solely on residue fragment analysis. We pioneer the development of the UniPTMs model, the first computational framework enabling full-site prediction across entire protein sequences. Using the latest UniProt database entries containing 136 residues, we implemented sliding window technology to systematically scan and visualize prediction outcomes (Figure 12). The model specifically targeted three modification types: O-glycosylated threonine (T), N-phosphorylated histidine (H), and succinylated lysine (K), with non-target site probabilities nullified and a confidence threshold set at 0.5. Experimental validation demonstrated 85% accuracy (17/20 target sites identified) and 83% sensitivity (5/6 positive sites detected), confirming the model's efficacy and reliability. This breakthrough establishes a novel paradigm for investigating protein functional regulation mechanisms through comprehensive PTM analysis.

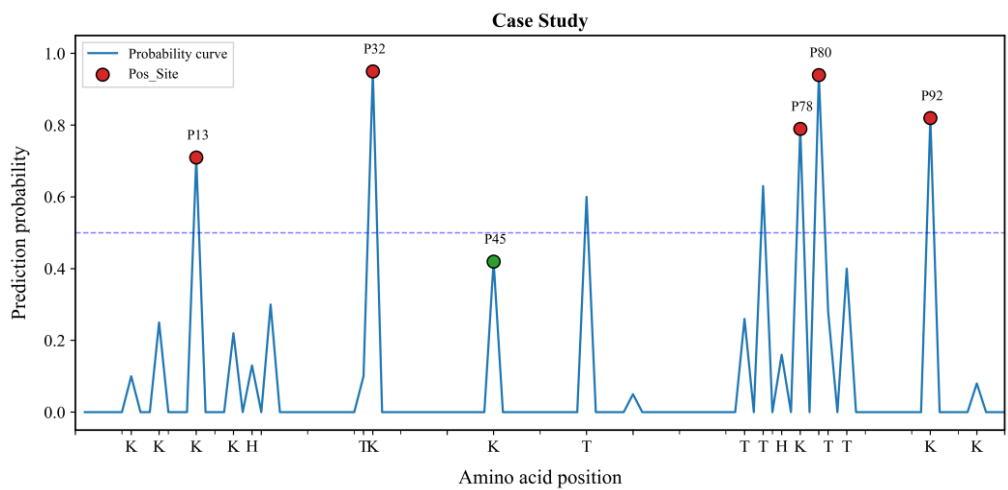

Figure 12. Predicted probability curves of histidine (H), lysine (K), and threonine (T) sites on a complete protein sequence. Positive sites are highlighted by red circles.

## 4. UniPTMs-mini

UniPTMs integrates all modules proposed in this study and demonstrates outstanding performance in practical applications. However, it exhibits relatively high model complexity. To achieve a more optimal balance between model complexity and performance, we have developed a lightweight version, UniPTMs-mini. This version judiciously streamlines partial fusion modules of UniPTMs and implements thorough optimizations of internal model dimensions. Although this adjustment resulted in a slight performance degradation, it significantly reduced the model complexity, thereby enhancing deployment efficiency while maintaining acceptable accuracy levels.

In the lightweight design of the UniPTMs-mini model, we implemented targeted architectural simplifications: the LDFN module was removed while retaining its core cross-dense attention mechanism to efficiently achieve feature fusion. For the BGCA module, we optimized the dimensional representation method of primary features, effectively reducing the dimensionality of fused features. This adjustment concurrently drove dimensionality reduction in related components of the BHGFN module. Furthermore, we systematically optimized the internal dimensions of other model components. At the training strategy level, we incorporated advanced techniques such as early stopping to further enhance training efficiency. Although these optimizations slightly compromised the original performance of UniPTMs-mini, they significantly reduced model complexity. To validate effectiveness, we conducted independent testing on UniPTMs-mini (as detailed in Table 15). Results demonstrate that UniPTMs-mini maintains leading performance levels, being only marginally inferior to our proposed UniPTMs model, successfully achieving an optimal balance between model complexity and performance.

Table 15. Independent test of UniPTMs-mini, where LOSE denotes the AP performance decrease percentage and UP represents the model inference speed improvement percentage

| DataSet | ACC | MCC | AUC | AP | LOSE | UP |
|---|---|---|---|---|---|---|
| T1 | 0.860 | 0.758 | 0.929 | 0.888 | 2.6% | 20% |
| T2 | 0.855 | 0.701 | 0.925 | 0.893 | 2.8% | 23% |
| H1 | 0.829 | 0.679 | 0.890 | 0.872 | 1.8% | 16% |
| H2 | 0.853 | 0.708 | 0.898 | 0.879 | 3% | 24% |
| K1 | 0.770 | 0.450 | 0.775 | 0.669 | 3.8% | 26% |
| K2 | 0.758 | 0.426 | 0.784 | 0.639 | 1.6% | 10% |
| C1 | 0.755 | 0.548 | 0.796 | 0.838 | 2% | 12% |
| C2 | 0.744 | 0.550 | 0.772 | 0.828 | 1.4% | 15% |

In summary, our proposed UniPTMs and UniPTMs-mini models provide flexible solutions to diverse research needs. For researchers pursuing ultimate performance, UniPTMs represents the ideal choice, while for those requiring lightweight models with maintained performance, UniPTMs-mini offers a more balanced solution. This dual-model framework effectively addresses the varying requirements of different research communities in terms of computational efficiency and predictive capability.

## 5.Discussion

This study innovatively proposes UniPTMs, the first unified framework for multi-PTM type prediction, which transcends the limitations of single modification prediction paradigms through a master-slave collaborative architecture enabling deep multimodal feature fusion. The framework implements a three-phase fusion strategy: The master branch utilizes ProtT5/ESM-2 embeddings as core features, establishing cross-modal interaction pathways via a Bidirectional Gated Cross-Attention (BGCA) module that integrates dynamic gating and multi-scale enhancement for profound semantic fusion. The slave branch combines EMBER2 low-dimensional structural embeddings with conventional features, employing a Low-Dimensional Fusion Network (LDFN) to achieve noise suppression and efficient interaction. A novel Multi-scale Adaptive Convolutional Pyramid (MACP) network is introduced to hierarchically extract multi-level sequence features, while a Bidirectional Hierarchical Gated Fusion Network (BHGFN) facilitates phased intermediate fusion between primary and auxiliary branches. Late-stage fusion adopts a Hierarchical Dynamic Weighting Fusion (HDWF) mechanism with triple-level weighting to reinforce primary-auxiliary relationships, incorporating biologically inspired dynamic regulation for adaptive integration. A hierarchical contrastive loss function is specifically designed to enhance feature consistency through intra-layer and cross-layer constraints, significantly improving predictive performance.The data and code are available at Github: https://github.com/Yuqiu-rgb/UniPTMs.

Although the UniPTMs framework has demonstrated remarkable advancements in predicting multi-type protein post-translational modification sites, this study reveals four critical dimensions requiring further improvement:

First, current and previous studies predominantly employ sequence-based protein pretrained language models (e.g., ESM-2, ProtT5) for representation learning. The recently released ESM-3 model, through scale expansion, data augmentation, architectural innovation, and multi-task co-optimization, has achieved comprehensive improvements over ESM-2 in protein representation capability, structural/functional prediction accuracy, and design efficiency, establishing itself as a benchmark achievement in protein language modeling. However, ESM-3 remains constrained by sequence conservation assumptions and inadequately captures three-dimensional structural information of proteins. With the emergence of structure-based deep learning models such as AlphaFold3, the development of sequence-structure joint embedding models has become feasible. AlphaFold3 explicitly encodes protein three-dimensional conformations, intermolecular interactions, and dynamic characteristics, providing direct physicochemical property correlations crucial for post-translational modification (PTM) site prediction, particularly demonstrating significant advantages in predicting structure-sensitive modifications like phosphorylation and glycosylation[56,57]. For instance, the MMFuncPhos model enhanced the AUC metric by 2.2% through integration of structural embedding features from AlphaFold2[32]. Future cross-modal representation approaches combining sequence and structural embeddings are anticipated to deliver more comprehensive and biologically interpretable information for PTM prediction.

Second, at the model architecture level, while the Bi-LSTM and Mamba hybrid architecture employed in this study demonstrates excellent performance in sequence modeling tasks, there remains room for optimization. Emerging architectures like Titans have broken through the context window limitations of traditional Transformer models by integrating Long Short-Term Memory (LSTM) networks with attention mechanisms[58]. This integration enables more efficient capture of long-range sequence dependencies while optimizing computational efficiency to address the modeling requirements of ultra-long protein sequences, thereby demonstrating superior performance competitiveness in temporal prediction and genomic tasks, and better alignment with feature extraction demands of novel sequence embeddings like ESM-3. However, in structural embedding feature extraction, traditional Convolutional Neural Networks (CNNs) struggle to effectively capture complex topological relationships in protein graph structures due to their lack of long-range dependency modeling capability and dynamic contextual awareness. In contrast, Graph Neural Networks (GNNs) inherently align with the non-Euclidean spatial architecture of proteins. Through direct modeling of atomic-level interactions, long-range dependencies, and physicochemical constraints, GNNs exhibit higher information fidelity and interpretability in structural embedding representations[59]. This characteristic grants them significant advantages in tasks requiring fine grained structural analysis, such as post-translational modification (PTM) site prediction. Future research directions should focus on synergistic integration of sequence modeling architectures like Titans with structural modeling frameworks such as GNNs, which holds promise for achieving breakthrough improvements in PTM prediction model performance.

Thirdly, while existing models continuously optimize prediction accuracy, their computational costs grow exponentially. For instance, replacing CNNs with GNNs leads to geometric progression in model parameters. To address this challenge, this study develops a lightweight version UniPTMs-mini through model compression techniques. However, this version exhibits an average accuracy drop of 2.35% compared to the original model while maintaining computational efficiency. This paradox highlights three critical research directions for future exploration: (1) Developing energy efficient hybrid architecture design strategies, such as achieving architectural lightweight through modular pruning and knowledge distillation; (2) Constructing adaptive feature selection mechanisms that dynamically screen critical structural and sequential features based on task requirements; (3) Designing dynamic computation allocation algorithms that adaptively adjust resource distribution according to input sample complexity, optimizing computational resource allocation without compromising prediction accuracy. The synergistic innovation of these technical pathways promises to overcome current performance efficiency trade-offs in PTM prediction models.

Finally, although the UniPTMs framework proposed in this study represents the first unified multi-type PTM prediction system that successfully breaks away from the conventional paradigm of single modification prediction, achieving significant methodological and conceptual innovations, its current application scope remains limited to five PTM types. This coverage falls substantially short of encompassing the extensive landscape of PTM diversity, leaving considerable exploration required to achieve accurate prediction for most modification types. Presently, existing research predominantly focuses on individual or limited PTM categories, exhibiting relatively constrained investigative perspectives[60]. However, the emergence of UniPTMs has pioneered a novel pathway in PTM prediction research. The integrated multi-type of prediction strategy it advocates, when synergistically combined with contemporary high performance model architectures, presents unprecedented developmental opportunities for this field. Under this trend, it is reasonable to anticipate that future advancements may enable comprehensive coverage and precise prediction of majority PTM types through unified modeling frameworks. Such progress would undoubtedly catalyze profound impacts and transformative developments in proteomics research and related disciplines.

## Funding

This research was funded by Natural Science Foundation of Jiangsu Province of China (Grant No. BK20230626), partly supported by the open funds of the State Key Laboratory of Plant Environmental Resilience (Grant No. SKLPERKF2401), supported by the Open project of State Key Laboratory of Animal Biotech Breeding (Grant No. 2024SKLAB6-1), the Fourth Batch of Leading Innovative Talents Introduction and Training Projects under the Longcheng Talent Plan in Changzhou City (Basic Research and Innovation) (Grant No. CQ20230086) and also supported by Changzhou Sci&Tech Program (Grant No. CJ20241083), the Development Project of Jilin Province of China (No. 20220508125RC).

## Ethical approval

This article does not contain any studies with animals performed by any of the authors.

## CRediT authorship contribution statement

**Yiyu Lin**: Conceptualization, Methodology. **Yan Wang**: Data curation. **You Zhou**: Writing-Original draft preparation, Software. **Xinye Ni**: Investigation. **Wujia Hui**: Data curation. **Sen Yang**: Visualization, Supervision.

## Declaration of competing interest

On behalf of all authors, the corresponding author states that there is no conflict of interest.